\documentclass[conference]{IEEEtran}
\IEEEoverridecommandlockouts
\usepackage{cite}
\usepackage{amsmath,amssymb,amsfonts}
\usepackage{algorithmic}
\usepackage{lipsum}
\usepackage{graphicx}
\usepackage{textcomp}
\usepackage{hyperref}
\usepackage{cleveref}
\usepackage[table,xcdraw]{xcolor}
\usepackage{booktabs, multirow} 
\usepackage{soul}
\usepackage[table]{xcolor} 
\usepackage{changepage,threeparttable} 
\usepackage{soul} 
\usepackage{enumitem} 

\def\BibTeX{{\rm B\kern-.05em{\sc i\kern-.025em b}\kern-.08em
    T\kern-.1667em\lower.7ex\hbox{E}\kern-.125emX}}

\usepackage{array}
\usepackage{multirow}
\usepackage{subcaption}
\usepackage{fancyhdr}  
\usepackage{lipsum}    

\newcommand{\imi}{\textbf{IMI}}
\newcommand{\kirby}{\textbf{KRPM}}

\newcommand{\icit}{\textbf{FARADAy}}
\newcommand{\icitmod}{\textbf{DIRAC}}
\newcommand{\icitfull}{\bful{F}lummoxing-\bful{A}malgam \bful{R}epresentation \bful{A}daptation with \bful{D}iscriminator as \bful{A}dversar\bful{Y}} 
\newcommand{\icitmodfull}{\bful{D}omain-\bful{I}ngrained \bful{R}epresentation with \bful{A}dversary as \bful{C}ritic}

\newcommand{\vbs}{\textbf{VanillaBSM}}
\newcommand{\vbm}{\textbf{VanillaBMM}}
\newcommand{\vbf}{\textbf{VanillaBFM}}

\newcommand{\bful}[1]{%
  \textbf{\underline{#1}}%
}

\begin{document}
\title{Adversarial Domain Adaptation for Metal Cutting Sound Detection: Leveraging Abundant Lab Data for Scarce Industry Data \\

\thanks{ 
}
}

\author{\IEEEauthorblockN{Mir imtiaz Mostafiz\textsuperscript{\textsection}}
\IEEEauthorblockA{\textit{Department of Computer Science} \\
\textit{Purdue University}\\
West Lafayette, USA \\
mmostafi@purdue.edu}
\and
\IEEEauthorblockN{Eunseob Kim\textsuperscript{\textsection}}
\IEEEauthorblockA{\textit{School of Mechanical Engineering} \\
\textit{Purdue University}\\
West Lafayette, USA \\
kim3235@purdue.edu}
\and
\IEEEauthorblockN{Adrian Shuai Li}
\IEEEauthorblockA{\textit{Department of Computer Science} \\
\textit{Purdue University}\\
West Lafayette, USA \\
li3944@purdue.edu}
\and
\IEEEauthorblockN{Elisa Bertino}
\IEEEauthorblockA{\textit{Department of Computer Science} \\
\textit{Purdue University}\\
West Lafayette, USA \\
bertino@purdue.edu}
\and
\IEEEauthorblockN{Martin Byung-Guk Jun}
\IEEEauthorblockA{\textit{School of Mechanical Engineering} \\
\textit{Purdue University}\\
West Lafayette, USA \\
mbgjun@purdue.edu}
\and
\IEEEauthorblockN{Ali Shakouri}
\IEEEauthorblockA{\textit{School of Electrical and Computer Engineering} \\
\textit{Purdue University}\\
West Lafayette, USA \\
shakouri@purdue.edu}
}

\maketitle

\begingroup\renewcommand\thefootnote{\textsection}
\footnotetext{Equal contribution to this work and designated as co-first authors.}
\endgroup
\thispagestyle{fancy}

\begin{abstract}
Cutting state monitoring in the milling process is crucial for improving manufacturing efficiency and tool life. Cutting sound detection using machine learning (ML) models, inspired by experienced machinists, can be employed as a cost-effective and non-intrusive monitoring method in a complex manufacturing environment. However, labeling industry data for training is costly and time-consuming. Moreover, industry data is often scarce. In this study, we propose a novel adversarial domain adaptation (DA) approach to leverage abundant lab data to learn from scarce industry data, both labeled, 
for training a cutting-sound detection model. 
Rather than adapting the features from separate domains directly, 
we project them first into two separate latent spaces that jointly work as the feature space for learning domain-independent representations. We also analyze two different mechanisms for adversarial learning where the discriminator works as an adversary and a critic in separate settings, enabling our model to learn expressive domain-invariant and domain-ingrained features, respectively. We collected cutting sound data from multiple sensors in different locations, prepared datasets from lab and industry domain, and evaluated our learning models on them. Experiments showed that our models outperformed the multi-layer perceptron based vanilla domain adaptation models in labeling tasks on the curated datasets, achieving near 92\%, 82\% and 85\%  accuracy respectively for three different sensors installed in industry settings. 

\end{abstract}



\section{Introduction}
Machine sound monitoring is a low-cost and non-intrusive method that converts acoustic signals into machine operational information. Manufacturing machines emit sounds that contain ample process information as they operate. However, the intricate interactions between components of manufacturing equipment and the complex nature of vibration and sound propagation make modeling sound emission challenging in the unpredictably noisy environment of a real-world factory floor~\cite{wegener2021noise}. In the metal cutting industry, continuous monitoring of the cutting state is crucial for evaluating process efficiency ~\cite{liu2018real} and estimating cutting tool life \cite{benkedjouh2015health}.

 Machine learning (ML) techniques have been applied to the detection of cutting sounds~\cite{kothuru2018application,chu2022research,peng2021sound,kim2023real-time,kim2023sound}. However, previous detectors have overlooked several critical factors, such as the numerous combinations of operational conditions, including machine type, tool, cutting parameters, and environmental factors. Moreover, the type and placement of sound sensors impact the prediction performance of ML models in machine sound monitoring~\cite{kim2023sound, kim2024operation}. In a controlled laboratory setting, collecting and labeling diverse cutting data for a Computer Numerical Control (CNC) machine tool can be automated~\cite{kim2024online}. However, in a real industrial setting, data collection and labeling are costly and time-consuming. This raises an interesting question: {\em Can lab data be useful for training a model that predicts real-world data?} An ML model simply trained on lab data is typically unable to make accurate predictions on real industry data due to distinct sound emission characteristics between different machines and settings. Therefore, to effectively train the model, it is crucial to consider the different domains created by varying machine types, cutting conditions, and sensor types and placements.

One approach to train an effective ML-based sound detector despite domain differences is through Domain Adaptation (DA)~\cite{ganin2016domain, tzeng2017adversarial, li2023building}. This method typically involves learning a common latent representation from two different domains, referred to as source and target domains. These domains are related but distinct; the source domain has a large training dataset, whereas the target domain has few labeled data obtained from the actual deployment environment of the equipment.
A classifier is then trained using such latent common representation, also referred to as {\em domain-independent representation} or {\em domain-invariant representation}. Because the representation is domain-invariant, the classifier can be ``transferred'' to make predictions in the target domain. For our research problem, the lab data is the labeled source domain and the industrial data is the target domain.

In this work, we collected the cutting sound datasets in both lab and industry settings with various sensor types and placements. 
Our work on designing a suitable DA approach to train a cutting detection model for deployment in a real factory floor, answers the following Research  Questions (RQs):

\newlist{researchquestions}{enumerate}{1}
\setlist[researchquestions,1]{
    label=RQ\arabic*),
    left=10pt,
    leftmargin=*,
    labelsep=0.5em,
    listparindent=1em,
    align=left
}

\begin{researchquestions}
    \item Which type of transformation for the sound features improves learning performances?
    \item How good is DA for learning on adequate source and inadequate target labeled data, and performing inference on unlabeled target domain data?
\end{researchquestions}
To answer the first question, we have experimentally evaluated single domain classification on both lab and industry data for separate sensors, after transforming their features in different ways. 
For the second research question, we compare two models: one has been developed by Li et al.~\cite{li2023building} but applied only the image data; the other is a modification of the first approach and our novel contribution. 

Our first model, developed by Li et.al. \cite{li2023building}, focuses on finding a common latent representation space for 
both domains, 
making the associated classifier learn in a domain-invariant way. To ensure that our model generates domain-invariant representations, we take an adversarial learning approach where the generator tries to fool a discriminator (whose job is to classify from which domain the sample is coming from). 
We refer the first model as \icit{} (\icitfull{})- as it fools (``flummoxes'') the adversarial discriminator by adapting through the mixing (``amalgamation'') of separate latent space representations for both domains. 

In our second model, 
we hypothesize that the classifier may need some domain information along with feature representations to infer the classes. So, unlike the first model, we try to generate a representation that can help the discriminator to properly classify the domains. If it is done properly, then it is ensured that the generator indeed has some bit of domain information ingrained in its generated latent representations- which the classifier can leverage. 
We refer to the second model as \icitmod{} (\icitmodfull{})- as in this methodology the generator will receive feedback from the discriminator more as a critic than as an adversary, to ingrain domain identity information in the shared representation space.

In \Cref{sec:relwork}, we provide background about ML-based sound monitoring techniques, and domain adaptation. Later, in \Cref{sec:datcol}, we discuss our process for lab and industry data collection and pre-processing for our DA task, 
Then, in \Cref{sec:method}, we formulate our research problem, and provide details about \icit{} and \icitmod{}. Finally, in \Cref{sec:exp}, we report the results of our evaluations with respect to our two research questions, by comparing different transformations and domain adaptation models. 

\section{Related Works} \label{sec:relwork}
\subsection{Machine sound monitoring technique}
To address noise from the surrounding environment, the Internal Sound Sensor (ISS) was developed, combining a stethoscope and a USB microphone~\cite{yun2020development}. Like a doctor using a stethoscope, the ISS focuses on target surface sounds while minimizing external interference, demonstrating better prediction accuracy for detecting manufacturing equipment operations compared to standard microphones~\cite{kim2024operation}. Industrial scenarios often present varied and limited types and locations of sound sensors, indicating a need for studies on different sensor domains, which have not yet been investigated. In machine sound monitoring, frequency domain techniques, particularly those based on Fast Fourier Transform (FFT), have been widely used for sound feature extraction and have proven effective as ML model inputs, demonstrating their ability in the industry ~\cite{kim2023sound,kim2023real-time,kim2024operation}. FFT captures the spectral content of the signal, highlighting dominant frequencies and harmonic structures~\cite{lauro2014monitoring}. 
A 1-D ML model using FFT features has shown the ability to detect subtle cutting sounds in micro-volume fine-milling processes and shown a fast response for real-time monitoring~\cite{kim2023real-time}. However, previous works have not explored the impact of FFT feature transformation on improving learning performance.

\subsection{Adversarial domain adaptation}

Adversarial Domain methods aim to learn a shared latent representation using two competing networks: a feature extractor and a  discriminator. One of the pioneering methods in this area is the Domain Adversarial Neural Network (DANN)~\cite{ganin2016domain}, which consists of three components: a feature extractor, a label predictor, and a domain classifier. In DANN, the generator is adversarially trained to maximize the domain classifier's loss by reversing its gradients. At the same time, the generator and label predictor work together to learn a representation that is domain-invariant. Similarly, the Adversarial Discriminative Domain Adaptation (ADDA)~\cite{tzeng2017adversarial} approach employs analogous network components but involves a multi-stage training process for these components. A recent method that aligns closely with our research is proposed by Li et al.~\cite{li2023building}. This approach combines DA with data augmentation to handle minimal and imbalanced training data from two domains, namely the source domain and the target domain, and to infer classes in target domains. However, these methods have primarily been applied and assessed in the context of image classification and object detection for source and target domains of images. Research has not explored DA methods for machine sound monitoring with the goal of training models for deployment across different machines, sensors, and environments.

\section{Data Collection and Dataset Preparation} \label{sec:datcol}

We collected operational sound data from CNC milling processes in both laboratory and industrial settings. 
\Cref{tab:datacollection} presents a comparison of the datasets. 
\begin{table}[bt]
\centering
\caption{Sound data collection comparison}\label{tab:datacollection}
\begin{tabular}{|cc|c|c|}
\hline
\multicolumn{2}{|c|}{Category} &
  Lab (\imi{}) &
  Industry (\kirby{}) \\ \hline
\multicolumn{2}{|c|}{\begin{tabular}[c]{@{}c@{}}Machine\\ (Model, Manuf.)\end{tabular}} &
  \begin{tabular}[c]{@{}c@{}}3-axis Vertical Mill\\ (VM20i, Hurco)\end{tabular} &
  \begin{tabular}[c]{@{}c@{}}4-axis Horizontal Mill\\ (HCN-6000, Mazak)\end{tabular} \\ \hline
\multicolumn{2}{|c|}{Working travel} &
  40$\times$20$\times$20 in. &
  31.5$\times$31.5$\times$31.5 in. \\ \hline
\multicolumn{1}{|c|}{\multirow{2}{*}{\begin{tabular}[c]{@{}c@{}}Max. spindle \\ spec.\end{tabular}}} &
  Torque &
  82.3 ft-lbs &
  442 ft-lbs \\ \cline{2-4} 
\multicolumn{1}{|c|}{} &
  Speed &
  12,000 rpm &
  10,000 rpm \\ \hline
\multicolumn{1}{|c|}{\multirow{3}{*}{\begin{tabular}[c]{@{}c@{}}\\Sensor\\ (Location)\end{tabular}}} &
  0 &
  \begin{tabular}[c]{@{}c@{}}Microphone\\ (Working area)\end{tabular} &
  \begin{tabular}[c]{@{}c@{}}Microphone\\ (Working area)\end{tabular} \\ \cline{2-4} 
\multicolumn{1}{|c|}{} &
  1 &
  \begin{tabular}[c]{@{}c@{}}ISS\\ (Near spindle)\end{tabular} &
  \begin{tabular}[c]{@{}c@{}}ISS\\ (Spindle)\end{tabular} \\ \cline{2-4} 
\multicolumn{1}{|c|}{} &
  2 &
  \begin{tabular}[c]{@{}c@{}}ISS\\ (Base)\end{tabular} &
  \begin{tabular}[c]{@{}c@{}}ISS\\ (Base)\end{tabular} \\ \hline
\multicolumn{2}{|c|}{Process} &
  End-milling &
  End-milling, facing, drilling \\ \hline
\multicolumn{2}{|c|}{Material} &
  Aluminium, Carbon steel &
  Cast iron \\ \hline
\multicolumn{2}{|c|}{\begin{tabular}[c]{@{}c@{}}Total data length\\ (Cutting time)\end{tabular}} &
  \begin{tabular}[c]{@{}c@{}}2789.97 sec.\\ (1032.97 sec.)\end{tabular} &
  \begin{tabular}[c]{@{}c@{}}315.38 sec.\\ (119.908 sec.)\end{tabular} \\ \hline
\end{tabular}
\end{table}The laboratory data, referred to as \textbf{IMI}, was gathered at the Indiana Next Generation Manufacturing Competitiveness Center (IN-MaC) of the Indiana Manufacturing Institute (IMI) at Purdue University. Detailed information on the laboratory data collection can be found 
in~\cite{kim2024online}. The industrial data, referred to as \textbf{KRPM}, was obtained from Kirby Risk Precision Machining (KRPM) in Indiana, a company specializing in the manufacturing of engine blocks for the heavy-duty industry. The raw sound data, recorded in the WAV format, along with comprehensive descriptions, are accessible at \url{https://github.com/purduelamm/mt_cutting_dataset}.
The two machines have different configurations, spindle capacities, and sizes. 
To accurately label the cutting sound, an expert reviewed recorded video footage of the operation while simultaneously listening to the sound from each sensor.
Two types of sound sensors were employed: (i) microphone, and (ii) ISS. Details of the ISS can be found in~\cite{yun2020development}. For each dataset three sound sensors were used to ensure a comprehensive data collection in the form of separate sound files. 
To create sound features, each sound file was loaded and transformed into the frequency domain using FFT with the Librosa library~\cite{mcfee2015librosa}. The sampling rate is 48 kHz and the FFT window was set to $\mathit{n_{fft} \text{ = 2,048}}$. Consequently, the FFT returned $\mathit{\text{1} + \frac{n_{fft}}{\text{2}}}=\text{1,025}$ frequency beans ranging from 0 Hz to 24 kHz. Each frequency bin value was squared to obtain the intensity, and then converted to Decibels (dB). Thus, each data sample has 1,025 frequency intensities in logarithmic scale as features. The number of samples in Librosa \cite{mcfee2015librosa} conversion is defined as $\frac{4 \times \text{sampling rate} \times \text{time} }{n_{fft}}$, resulting in 261,560 samples for \imi{} and 12,213 samples for \kirby{}. The number of cut (non-cut) sample frames were 97,312 (164,248) and 5,405 (6,805), for \imi{} and \kirby{}, respectively.

\section{Methodology} \label{sec:method}

Our approach is based on the methodology by Li et al.~\cite{li2023building}, which uses DA along with data augmentation for training a target model from minimal and imbalanced training data from two domains (e.g., source and target). This model is able to infer classes for samples from the target domains. We first give our problem statement in \Cref{sec:probstate}, followed by a description of the \icit{} model in \Cref{sec:icit}, and our new approach \icitmod{} in \Cref{sec:icitmod}.

\subsection{Problem Statement} \label{sec:probstate}
\textbf{In words}, our research problem can be stated as follows: 
\vspace{0.5\baselineskip}

\emph{``Given a large labeled dataset from a source domain, and a way smaller labeled dataset from a target domain, can we train a model which can classify samples
from the target domain, by learning domain dependent and domain independent features from source and target domains?''}\\

Formally, {let $\mathcal{D}_s = \{(\mathbf{x}_s^{(i)}, y_s^{(i)})\}_{i=1}^{n_s}$, be a labeled dataset of size $n_s$, where sample data $\mathbf{x}_s^{(i)} \in \mathbf{S}$ -- the source domain. Also, let another labelled dataset be $\mathcal{D}_t= \{(\mathbf{x}_t^{(i)}, y_t^{(i)})\}_{i=1}^{n_t}$, of size $n_t$, where sample data $\mathbf{x}_t^{(i)} \in \mathbf{T}$ -- the target domain. Moreover, $y_s^{(i)}, y_t^{(i)} \in \{0,1\}$ are class labels for each sample irrespective of their domain. It is assumed that $n_s \gg n_t$.} Let $f_{dep}(.; \phi)$ \& $f_{indep}(.;\psi)$ be two models which learn domain dependent and independent features respectively ($\phi, \psi$ are learnable parameters). In other words, for any domain 
$\mathbf{\zeta}$ $\in \{\mathbf{S}, \mathbf{T}\}$, $ \forall_{\mathbf{x} \in \zeta } f_{indep}(\mathbf{x}| \mathbf{x} \in \mathbf{S};\psi ) = f_{indep}(\mathbf{x}| \mathbf{x} \in \mathbf{T};\psi )$ \& $ \exists_{\mathbf{x} \in \zeta } f_{dep}(\mathbf{x}| \mathbf{x} \in \mathbf{S};\phi ) \neq f_{dep}(\mathbf{x}| \mathbf{x} \in \mathbf{T};\phi )$. Let $f(.; \theta)$ be a binary-classification model with learnable parameters $\theta$. For brevity, we denote $f_{indep}(\mathbf{x};\psi )$ \& $f_{dep}(\mathbf{x}|\mathbf{x} \in \zeta;\phi )$ as $f_{\psi}(\mathbf{x})$ \& $f_{\phi}(\mathbf{x}, \mathbf{\zeta})$. \textbf{In terms of mathematical notations}, our problem can be stated as: 
\vspace{0.5\baselineskip}

\emph{``Can we learn a classification model $f(.; \theta)$, and feature learning models $f_{\psi}(.), f_{\phi}(.)$ from labeled dataset $\mathcal{D}_s$ of size $n_s$ from source domain $\mathbf{S}$ and labeled dataset $\mathcal{D}_t$ of size $n_t$ from target domain $\mathbf{T}$ (where $n_s \gg n_t$), such that for a target data sample $\mathbf{x_t} \in \mathbf{T}$, $f( \mathbf{x_t}, f_{\psi}(\mathbf{x_t}) , f_{\phi}(\mathbf{x_t},\mathbf{T}); \theta)$ can predict the associated class label $\mathbf{y_t} \in \{0,1\}$?''}.

\subsection{\icit{}- Flummoxing-Amalgam Representation Adaptation with Discriminator as Adversary} \label{sec:icit}
\begin{figure}[b]
\centerline{\includegraphics[scale=0.37]{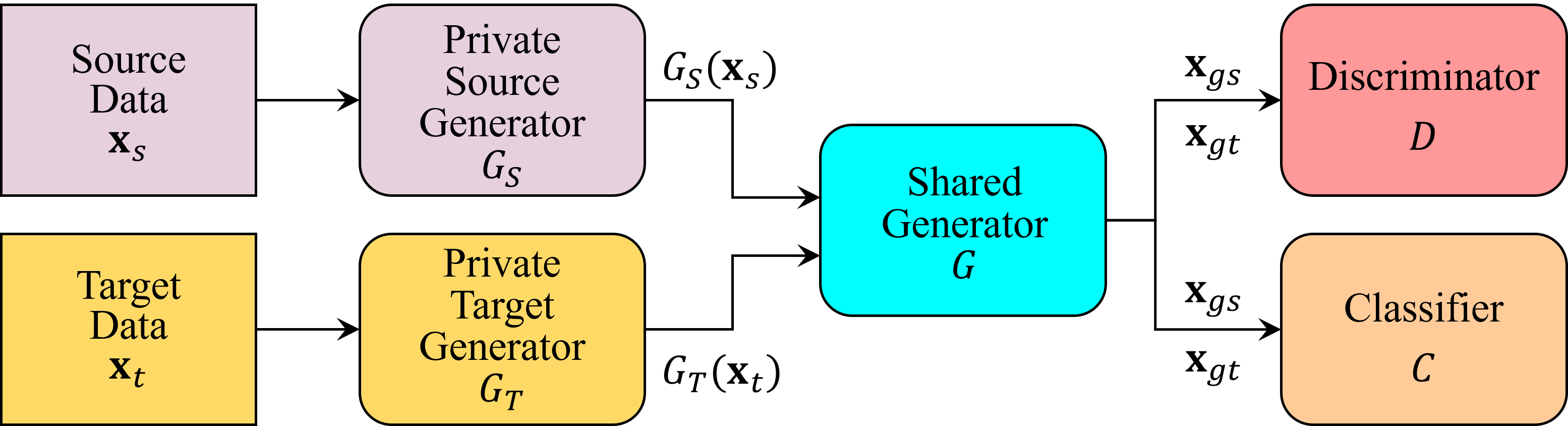}}
\caption{\icit{} Architecture from~\cite{li2023building}}\label{fig:icit}
\end{figure}

Similar to other adversarial training frameworks, \icit{} also consists of generator(s) and a discriminator (see Figure~\ref{fig:icit}).
The goal of the adversarial training here is to generate features so that the discriminator cannot distinguish from which domain the features are. The model has two \textit{private} generators for source and target domains separately, $G_S$ and $G_T$, respectively. The corresponding generator takes input sample from a domain and is trained to generate private or domain-dependent features. It also has a \textit{shared} generator $G$ that takes those features as input, for both source and target domain, and generates domain-independent features. This generator directly tries to fool the discriminator $D$ and hence gets feedback in the form of loss propagation to learn how to generate domain-independent features. Practically, the \textit{shared} generator ensures that the classifier $C$ does not require domain information for inference; it only needs to obtain shared feature representations. In a sense, the classifier $C$ learns exclusively from domain-independent features, as the domain-dependent ones are transformed by the generator. 

Formally, in the forward pass, for source domain sample $(\mathbf{x}_s, y_s)$ and target domain sample $(\mathbf{x}_t, y_t)$, we pass them to \textit{Private Source Generator}, $G_S$ and \textit{Private Target Generator}, $G_T$, respectively. We get private features $G_S(\mathbf{x}_s)$ and $G_T(\mathbf{x}_t)$, respectively. 
Then, we pass them to \textit{Shared Generator}, \( G \), to obtain the shared representations, 
\begin{align} \mathbf{x}_{gs} = G(G_S(\mathbf{x}_s)), \quad \mathbf{x}_{gt} = G(G_T(\mathbf{x}_t)) \end{align}
Then, we pass them to \textit{Discriminator}, \( D \) to obtain
\begin{align} \hat{\mathbf{d}}_{s} = D(\mathbf{x}_{gs}), \quad \hat{\mathbf{d}}_{t} = D(\mathbf{x}_{gt}) \end{align}
and also to \textit{Classifier}, $C$ to obtain \begin{align} \hat{\mathbf{y}}_{s} = C(\mathbf{x}_{gs}), \quad \hat{\mathbf{y}}_{t} = C(\mathbf{x}_{gt}) \end{align}

In backward pass, we compute three losses: (i) classification loss- which is a cross-entropy loss,  \begin{align}\mathcal{L}_c = -{\mathbf{y}_{s}} \cdot \log(\hat{\mathbf{y}_{s}}) - \lambda {\mathbf{y}_{t}} \cdot \log(\hat{\mathbf{y}_{t}})\end{align}(ii) discriminator loss- similar to cross-entropy as the discriminator itself is a classifier for domains,\begin{align}\mathcal{L}_d = -{\mathbf{d}_{s}} \cdot \log(\hat{\mathbf{d}_{s}}) - {\mathbf{d}_{t}} \cdot \log(\hat{\mathbf{d}_{t}})\end{align}  and (iii) Generator loss- it has the inverse form of discriminator loss so that when the discriminator predicts correctly (incorrectly), it gets a negative (positive) reinforcement, \begin{align}\mathcal{L}_g = -{(\mathbf{1}-\mathbf{d}_{s})} \cdot \log(\hat{\mathbf{d}_{s}}) - {(\mathbf{1}-\mathbf{d}_{t})} \cdot \log(\hat{\mathbf{d}_{t}})\end{align} 
 In the backpropagation step, 
classifier and the discriminator are only updated by their respective loss gradients, and the generators, receive feedback from both. So, the gradients are \begin{align}\nabla C = \beta \frac{\partial \mathcal{L}_C}{\partial {C}}, \nabla D = \beta \frac{\partial \mathcal{L}_D}{\partial {D}}, \nabla G = \beta \frac{\partial \mathcal{L}_g}{\partial G} + \gamma \frac{\partial \mathcal{L}_c}{\partial G}\end{align} \begin{align}\nabla G_{S} = \beta \frac{\partial \mathcal{L}_g}{\partial G_S} + \gamma \frac{\partial \mathcal{L}_c}{\partial G_S}, \nabla G_{T} = \beta \frac{\partial \mathcal{L}_g}{\partial G_T} + \gamma \frac{\partial \mathcal{L}_c}{\partial G_T}\end{align}Here, $\lambda, \beta, \gamma$ are the hyperparameters. 

\subsection{\icitmod{}- Domain-Ingrained Representation with Adversary as Critic
} \label{sec:icitmod} 
\icitmod{} is a variant of \icit{} where we modified the model training part based on some intuitions. Firstly, we hypothesize that when the generator $G$ receives private features, the information about the domain from which the features originate is not passed to $G$. 
However, the Classifier $C$ might need some domain-specific features. Therefore, if $G$ can have some latent information about the domain from which the generated features originate, it can generate some domain-dependent features to help the classifier. For this, the discriminator $D$ can help $G$. It will try to correctly classify the domains, and will feedback the loss accordingly to the generator; if $D$ classifies the domain correctly (incorrectly), $G$ will get positive (negative) reinforcement. This ensures that $G$ learns some domain-dependent features too. Hence, the generator loss is now the same as the discriminator loss: \begin{align}\mathcal{L}_g = \mathcal{L}_d = -{\mathbf{d}_{s}} \cdot \log(\hat{\mathbf{d}_{s}}) - {\mathbf{d}_{t}} \cdot \log(\hat{\mathbf{d}_{t}})\end{align} 

Secondly, in \icit{}, source and target classification losses are combined before being passed to the private generators $G_S$ and $G_T$. 
But they should only be responsible for their own domain losses. Hence, we split the classification losses into two components: source and target classification losses:
\begin{equation}
\mathcal{L}_{cs}= -{\mathbf{y}_{s}} \cdot \log(\hat{\mathbf{y}}_{s}), \quad \mathcal{L}_{ct}= -{\mathbf{y}}_{t} \cdot \log(\hat{\mathbf{y}}_{t})
\end{equation}
The classification loss is: \begin{align}\mathcal{L}_c = \mathcal{L}_{cs} + \lambda \mathcal{L}_{ct}\end{align}
Now, the gradient for \textit{private generators} are \begin{align}\nabla G_{S} = \beta \frac{\partial \mathcal{L}_g}{\partial G_S} + \gamma \frac{\partial \mathcal{L}_{cs}}{\partial G_S}, \nabla G_{T} = \beta \frac{\partial \mathcal{L}_g}{\partial G_T} + \gamma \frac{\partial \mathcal{L}_{ct}}{\partial G_T}\end{align}

Lastly, instead of a single pass to update the discriminator, classifier, and generators together, we follow the training techniques by Goodfellow et al.~\cite{goodfellow2020generative}. 
For each training step, we take some forward 
passes and update only the discriminator $D$ in each pass. After that, we execute 
again a single forward pass, and update $G$, $G_S$, $G_T$, and $C$. This helps to prevent overfitting and enhance faster convergence.

\subsection{Model Architecture} \label{sec:mod:arch}

All private generators $G_S, G_T$ have \textbf{Multi Layer Perceptron (MLP)} architecture with 1,025 input dimension, two hidden layers with 768 and 512 nodes, with Sigmoid activations. The shared generator $G$ is also an \textbf{MLP} with 512 input dimensions and a hidden layer with 200 nodes with ReLU activations. Both the discriminator $D$ and the classifier $C$ are also \textbf{MLP} with 200 input dimension and a dense layer of 100 nodes with ReLU activation. For output layer, Softmax is applied. All layers are optimized using the ADAM optimizer \cite{kingma2014adam} with a learning rate of $2 \times 10 ^ {-4}$ and a beta decay of 0.02.

\section{Experiments} \label{sec:exp}

\subsection{Experiment settings}
We implemented all our models using 
TensorFlow~\cite{tensorflow2015-whitepaper}, and Librosa~\cite{mcfee2015librosa}. The models were run on an NVIDIA GeForce RTX 3090 GPU with Cuda 12.2 support. For experiments, we randomly split our datasets in (70-10-20)\% ratio for train, validation, and test set for each sensor. 
\subsection{RQ1 Experiment}
The motivation behind our RQ1 experiment is to determine if we need to transform the source and target domain data before passing them to the model. Our hypothesis is that the feature distributions of the source and target domains differ. If a prior transformation can improve single-domain classification performance without adding more deep layers, we can use these transformed features directly for transfer learning. This approach helps us avoid introducing more layers, thus learning fewer parameters and optimizing the transfer learning process.
For this, we introduce a simple binary classification model comprising of a \textbf{MLP} with a single hidden dense layer with 512 neurons. We use sigmoid and softmax as the activation functions in the hidden and output layers respectively. We ran the model for 20 epochs, with the batch size set as 128. We also used the same splits as before, and reported our results on the validation set instead of the test set, as the validation set results reflect more on the model’s generalizablity. Better accuracy in validation ensures that these transformations will also work better for transfer learning task. We loaded all \imi{} data and ran the model. The model was simpler and hence it did not take much memory, thereby we could load the whole data in memory. The evaluated transformations are: (i) Identity ($idx$), (ii) Standardization ($stanx$), (iii) Log ($logx$), (iv) Sigmoid ($sigmoidx$) and, (v) Gamma ($gammax$). In the second column of \Cref{tab:rq1_merged2}, we have described the transformation formulas for the feature value $x$.  



From \Cref{tab:rq1_merged2}, we can see that for all sensors of both datasets, $gammax$ transformation outperforms all others in validation accuracy, thus being the representation with the greater generalization ability. $stanx$ has the best performance for training accuracy in \imi{} ($gammax$ is not very far). Overall, $gammax$ with $\gamma=4.2$ proves to be the best choice for data pre-processing transformation. 
\textbf{In RQ2 experiments, models are trained on Gamma-transformed features}.
\begin{table}[tb]
\caption{RQ1- Performance Comparison of Different Features Transformations on each sensor}\label{tab:rq1_merged2}
\centering
\resizebox{\columnwidth}{!}{%
\begin{tabular}{|c|c|ccc|ccc|}
\hline
\multirow{2}{*}{\begin{tabular}[c]{@{}c@{}}\textbf{Trans-}\\ \textbf{formation}\end{tabular}} 
&
\multirow{2}{*}{\begin{tabular}[c]{@{}c@{}}\textbf{Formula}\end{tabular}} &
  \multicolumn{3}{c|}
  { \begin{tabular}[c]{@{}c@{}}\imi{} Train (Validation)\\  Accuracy in \%\end{tabular}} &
  \multicolumn{3}{c|}{\begin{tabular}[c]{@{}c@{}}\kirby{} Train (Validation)\\  Accuracy in \%\end{tabular}}
  \\ 
  \cline{3-8} 
 &
 &
  \multicolumn{1}{c|}{Sensor 0} &
  \multicolumn{1}{c|}{Sensor 1} &
  \multicolumn{1}{c|}{Sensor 2} 
  &

  \multicolumn{1}{c|}{Sensor 0} &
  \multicolumn{1}{c|}{Sensor 1}
  &
  \multicolumn{1}{c|}{Sensor 2} 
  
  \\ \hline
$idx$ & $x$ &
  \multicolumn{1}{c|}{\begin{tabular}[c]{@{}c@{}}69.6\\ (62.6)\end{tabular}} &
  \multicolumn{1}{c|}{\begin{tabular}[c]{@{}c@{}}69.7\\ (62.8)\end{tabular}} &
  \begin{tabular}[c]{@{}c@{}}62.8\\ (62.3)\end{tabular} &
  \multicolumn{1}{c|}{\begin{tabular}[c]{@{}c@{}}53.7\\ (55.1)\end{tabular}} &
  \multicolumn{1}{c|}{\begin{tabular}[c]{@{}c@{}}52.3\\ (55.8)\end{tabular}} &
  \begin{tabular}[c]{@{}c@{}}53.4\\ (54.4)\end{tabular} \\ \hline
$stanx$ & $\frac{x - \text{min}(x)}{\text{max}(x) - \text{min}(x)}$ &
  \multicolumn{1}{c|}{\begin{tabular}[c]{@{}c@{}}\textbf{98.7}\\ (99.2)\end{tabular}} &
  \multicolumn{1}{c|}{\begin{tabular}[c]{@{}c@{}}\textbf{98.4}\\ (99.3)\end{tabular}} &
  \begin{tabular}[c]{@{}c@{}}\textbf{99.3}\\ (99.2)\end{tabular} &
  \multicolumn{1}{c|}{\begin{tabular}[c]{@{}c@{}}91.7\\ (78.6)\end{tabular}} &
  \multicolumn{1}{c|}{\begin{tabular}[c]{@{}c@{}}91.6\\ (83.6)\end{tabular}} &
  \begin{tabular}[c]{@{}c@{}}80.3\\ (54.4)\end{tabular} \\ \hline
$logx$ & $(\frac{255 \text{log}(81 + x)}{\text{log}(81 + \text{max}(x))})$ &
  \multicolumn{1}{c|}{\begin{tabular}[c]{@{}c@{}}67.6\\ (94.2)\end{tabular}} &
  \multicolumn{1}{c|}{\begin{tabular}[c]{@{}c@{}}67.7\\ (97.2)\end{tabular}} &
  \begin{tabular}[c]{@{}c@{}}94.1\\ (97.2)\end{tabular} &
  \multicolumn{1}{c|}{\begin{tabular}[c]{@{}c@{}}54.0\\ (54.1)\end{tabular}} &
  \multicolumn{1}{c|}{\begin{tabular}[c]{@{}c@{}}52.7\\ (73.3)\end{tabular}} &
  \begin{tabular}[c]{@{}c@{}}53.8\\ (72.7)\end{tabular} \\ \hline
$sigmoidx$ & $\frac{255e^{x+41}}{1+ e^{x+41}}$ &
  \multicolumn{1}{c|}{\begin{tabular}[c]{@{}c@{}}95.1\\ (99.0)\end{tabular}} &
  \multicolumn{1}{c|}{\begin{tabular}[c]{@{}c@{}}95.0\\ (99.2)\end{tabular}} &
  \begin{tabular}[c]{@{}c@{}}99.0\\ (99.1)\end{tabular} &
  \multicolumn{1}{c|}{\begin{tabular}[c]{@{}c@{}}89.1\\ (82.0)\end{tabular}} &
  \multicolumn{1}{c|}{\begin{tabular}[c]{@{}c@{}}86.0\\ (94.1)\end{tabular}} &
  \begin{tabular}[c]{@{}c@{}}93.2\\ (82.1)\end{tabular} \\ \hline
$gammax$ & $(\frac{255(81+x)}{81+\max{(x)}})^\gamma$ &
  \multicolumn{1}{c|}{\begin{tabular}[c]{@{}c@{}}98.4\\ (\textbf{99.3})\end{tabular}} &
  \multicolumn{1}{c|}{\begin{tabular}[c]{@{}c@{}}98.3\\ (\textbf{99.4})\end{tabular}} &
  \begin{tabular}[c]{@{}c@{}}\textbf{99.3}\\ (\textbf{99.3})\end{tabular} &
  \multicolumn{1}{c|}{\begin{tabular}[c]{@{}c@{}}\textbf{100.0}\\ (\textbf{91.4})\end{tabular}} &
  \multicolumn{1}{c|}{\begin{tabular}[c]{@{}c@{}}\textbf{95.9}\\ (\textbf{95.2})\end{tabular}} &
  \begin{tabular}[c]{@{}c@{}}\textbf{95.0}\\ (\textbf{88.3})\end{tabular} \\ \hline
\end{tabular}}
\end{table}




\begin{table*}[htbp]
\centering
\caption{RQ2- Target Test Accuracy (in \%) Comparison among transfer learning models with \imi{} Sensors as Source, \& \kirby{} sensors as Target, with varying data size}
\label{tab:rq2sensall_reduced_new}
\begin{tabular}{|c|c|ccccccccccccccc|}
\hline \hline 
\multicolumn{17}{c}{Source: \imi{} Sensor \textbf{0}} \\\hline
\hline

\multicolumn{2}{|c|}{\begin{tabular}[c]{@{}c@{}}\\ Target Sensor\\ Index $\rightarrow$\end{tabular}} &
\multicolumn{5}{c|}{Sensor 0} &
  \multicolumn{5}{c|}{Sensor 1} &
  \multicolumn{5}{c|}{Sensor 2} 
  
  \\ \cline{1-17} 
  
\multicolumn{2}{|c|}{\begin{tabular}[c]{@{}c@{}}\\ Model \\ $\rightarrow$\end{tabular}} &
  \multicolumn{1}{c}{\multirow{2}{*}{\begin{tabular}[c]{@{}c@{}}\\ \textbf{Vanilla}\\ \textbf{BSM}\end{tabular}}} &
  \multicolumn{1}{c}{\multirow{2}{*}{\begin{tabular}[c]{@{}c@{}}\\ \textbf{Vanilla}\\ \textbf{BMM}\end{tabular}}} &
  \multicolumn{1}{c}{\multirow{2}{*}{\begin{tabular}[c]{@{}c@{}}\\ \textbf{Vanilla}\\ \textbf{BFM}\end{tabular}}} &
  \multicolumn{1}{c}{\multirow{2}{*}{\begin{tabular}[c]{@{}c@{}}\\ \textbf{FARADAy}\end{tabular}}} &
  \multicolumn{1}{c|}{\multirow{2}{*}{\begin{tabular}[c]{@{}c@{}}\\ \textbf{DIRAC}\end{tabular}}} &
  \multicolumn{1}{c}{\multirow{2}{*}{\begin{tabular}[c]{@{}c@{}}\\ \textbf{Vanilla}\\ \textbf{BSM}\end{tabular}}} &
  \multicolumn{1}{c}{\multirow{2}{*}{\begin{tabular}[c]{@{}c@{}}\\ \textbf{Vanilla}\\ \textbf{BMM}\end{tabular}}} &
  \multicolumn{1}{c}{\multirow{2}{*}{\begin{tabular}[c]{@{}c@{}}\\ \textbf{Vanilla}\\ \textbf{BFM}\end{tabular}}} &
  \multicolumn{1}{c}{\multirow{2}{*}{\begin{tabular}[c]{@{}c@{}}\\ \textbf{FARADAy}\end{tabular}}} &
  \multicolumn{1}{c|}{\multirow{2}{*}{\begin{tabular}[c]{@{}c@{}}\\ \textbf{DIRAC}\end{tabular}}}
  &
  \multicolumn{1}{c}{\multirow{2}{*}{\begin{tabular}[c]{@{}c@{}}\\ \textbf{Vanilla}\\ \textbf{BSM}\end{tabular}}} &
  \multicolumn{1}{c}{\multirow{2}{*}{\begin{tabular}[c]{@{}c@{}}\\ \textbf{Vanilla}\\ \textbf{BMM}\end{tabular}}} &
  \multicolumn{1}{c}{\multirow{2}{*}{\begin{tabular}[c]{@{}c@{}}\\ \textbf{Vanilla}\\ \textbf{BFM}\end{tabular}}} &
  \multicolumn{1}{c}{\multirow{2}{*}{\begin{tabular}[c]{@{}c@{}}\\ \textbf{FARADAy}\end{tabular}}} &
  \multicolumn{1}{c|}{\multirow{2}{*}{\begin{tabular}[c]{@{}c@{}}\\ \textbf{DIRAC}\end{tabular}}}

  \\ 
  \cline{1-2} 
  {\begin{tabular}[c]{@{}c@{}}\\ Source\\Labeled\\Data\\Size $\downarrow$\\\  \end{tabular}}
 &  {\begin{tabular}[c]{@{}c@{}}\\ Target\\Labeled\\Data\\Size$\downarrow$\\\  \end{tabular}}
   & \multicolumn{1}{c}{}& \multicolumn{1}{c}{} & \multicolumn{1}{c}{} & \multicolumn{1}{c}{} & \multicolumn{1}{c|}{} & \multicolumn{1}{c}{} & \multicolumn{1}{c}{} & \multicolumn{1}{c}{} &
   \multicolumn{1}{c}{} & \multicolumn{1}{c|}{} & \multicolumn{1}{c}{} & \multicolumn{1}{c}{} & \multicolumn{1}{c}{} & \multicolumn{1}{c}{} &
   \multicolumn{1}{c|}{}
   
   \\ \hline 
\multirow{4}{*}{2,000} &
  50 &
  \multicolumn{1}{c}{45.9} &
  \multicolumn{1}{c}{44.6} &
  \multicolumn{1}{c}{51.3} &
  \multicolumn{1}{c}{77.3} &
  \multicolumn{1}{c|}{\textbf{79.1}} &
  \multicolumn{1}{c}{43.7} &
  \multicolumn{1}{c}{45.8} &
  \multicolumn{1}{c}{49.7} &
  \multicolumn{1}{c}{\textbf{72.8}} &
  \multicolumn{1}{c|}{72.5} &
  \multicolumn{1}{c}{43.3} &
  \multicolumn{1}{c}{44.0} &
  \multicolumn{1}{c}{46.9} &
  \multicolumn{1}{c}{73.4} &
  \textbf{74.2} \\ 
 &
  500 &
  \multicolumn{1}{c}{44.7} &
  \multicolumn{1}{c}{45.8} &
  \multicolumn{1}{c}{50.1} &
  \multicolumn{1}{c}{\textbf{88.1}} &
  \multicolumn{1}{c|}{87.4} &
  \multicolumn{1}{c}{43.7} &
  \multicolumn{1}{c}{47.2} &
  \multicolumn{1}{c}{53.4} &
  \multicolumn{1}{c}{77.0} &
  \multicolumn{1}{c|}{\textbf{80.5}} &
  \multicolumn{1}{c}{43.7} &
  \multicolumn{1}{c}{49.2} &
  \multicolumn{1}{c}{48.6} &
  \multicolumn{1}{c}{79.5} &
  \textbf{80.2} \\ 
 &
  4,000 &
  \multicolumn{1}{c}{45.1} &
  \multicolumn{1}{c}{50.1} &
  \multicolumn{1}{c}{56.5} &
  \multicolumn{1}{c}{\textbf{91.3}} &
  \multicolumn{1}{c|}{89.0} &
  \multicolumn{1}{c}{43.3} &
  \multicolumn{1}{c}{54.3} &
  \multicolumn{1}{c}{56.3} &
  \multicolumn{1}{c}{80.8} &
  \multicolumn{1}{c|}{\textbf{81.9}} &
  \multicolumn{1}{c}{43.7} &
  \multicolumn{1}{c}{48.4} &
  \multicolumn{1}{c}{56.3} &
  \multicolumn{1}{c}{80.7} &
  \textbf{82.8} \\ 
 &
  12,213 &
  \multicolumn{1}{c}{45.5} &
  \multicolumn{1}{c}{56.3} &
  \multicolumn{1}{c}{56.3} &
  \multicolumn{1}{c}{\textbf{92.5}} &
  \multicolumn{1}{c|}{91.6} &
  \multicolumn{1}{c}{43.7} &
  \multicolumn{1}{c}{56.3} &
  \multicolumn{1}{c}{56.3} &
  \multicolumn{1}{c}{80.7} &
  \multicolumn{1}{c|}{\textbf{82.9}} &
  \multicolumn{1}{c}{43.7} &
  \multicolumn{1}{c}{56.3} &
  \multicolumn{1}{c}{56.3} &
  \multicolumn{1}{c}{81.1} &
  \textbf{84.5} \\ \hline
\multirow{4}{*}{8,000} &
  50 &
  \multicolumn{1}{c}{44.5} &
  \multicolumn{1}{c}{40.7} &
  \multicolumn{1}{c}{46.1} &
  \multicolumn{1}{c}{78.8} &
  \multicolumn{1}{c|}{\textbf{79.8}} &
  \multicolumn{1}{c}{44.1} &
  \multicolumn{1}{c}{43.9} &
  \multicolumn{1}{c}{46.6} &
  \multicolumn{1}{c}{71.4} &
  \multicolumn{1}{c|}{\textbf{73.8}} &
  \multicolumn{1}{c}{43.7} &
  \multicolumn{1}{c}{43.8} &
  \multicolumn{1}{c}{48.5} &
  \multicolumn{1}{c}{\textbf{73.1}} &
  72.0 \\ 
 &
  500 &
  \multicolumn{1}{c}{45.2} &
  \multicolumn{1}{c}{48.0} &
  \multicolumn{1}{c}{49.7} &
  \multicolumn{1}{c}{87.1} &
  \multicolumn{1}{c|}{\textbf{89.2}} &
  \multicolumn{1}{c}{44.1} &
  \multicolumn{1}{c}{44.0} &
  \multicolumn{1}{c}{53.6} &
  \multicolumn{1}{c}{80.1} &
  \multicolumn{1}{c|}{\textbf{80.5}} &
  \multicolumn{1}{c}{43.7} &
  \multicolumn{1}{c}{48.0} &
  \multicolumn{1}{c}{51.3} &
  \multicolumn{1}{c}{81.8} &
  \textbf{82.1} \\ 
 &
  4,000 &
  \multicolumn{1}{c}{44.2} &
  \multicolumn{1}{c}{47.8} &
  \multicolumn{1}{c}{56.5} &
  \multicolumn{1}{c}{\textbf{92.4}} &
  \multicolumn{1}{c|}{91.2} &
  \multicolumn{1}{c}{44.0} &
  \multicolumn{1}{c}{52.7} &
  \multicolumn{1}{c}{56.3} &
  \multicolumn{1}{c}{81.6} &
  \multicolumn{1}{c|}{\textbf{83.2}} &
  \multicolumn{1}{c}{43.7} &
  \multicolumn{1}{c}{52.2} &
  \multicolumn{1}{c}{56.3} &
  \multicolumn{1}{c}{83.1} &
  \textbf{84.2} \\ 
 &
  12,213 &
  \multicolumn{1}{c}{43.4} &
  \multicolumn{1}{c}{53.7} &
  \multicolumn{1}{c}{56.3} &
  \multicolumn{1}{c}{\textbf{92.5}} &
  \multicolumn{1}{c|}{91.9} &
  \multicolumn{1}{c}{46.0} &
  \multicolumn{1}{c}{48.8} &
  \multicolumn{1}{c}{56.3} &
  \multicolumn{1}{c}{80.2} &
  \multicolumn{1}{c|}{\textbf{82.9}} &
  \multicolumn{1}{c}{43.7} &
  \multicolumn{1}{c}{56.3} &
  \multicolumn{1}{c}{56.3} &
  \multicolumn{1}{c}{80.2} &
  \textbf{84.9} \\ \hline
\multirow{4}{*}{16,000} &
  50 &
  \multicolumn{1}{c}{45.0} &
  \multicolumn{1}{c}{41.8} &
  \multicolumn{1}{c}{50.5} &
  \multicolumn{1}{c}{79.7} &
  \multicolumn{1}{c|}{\textbf{79.8}} &
  \multicolumn{1}{c}{44.1} &
  \multicolumn{1}{c}{44.1} &
  \multicolumn{1}{c}{44.4} &
  \multicolumn{1}{c}{\textbf{73.5}} &
  \multicolumn{1}{c|}{72.3} &
  \multicolumn{1}{c}{44.0} &
  \multicolumn{1}{c}{44.8} &
  \multicolumn{1}{c}{51.3} &
  \multicolumn{1}{c}{73.6} &
  \textbf{73.9} \\ 
&
  500 &
  \multicolumn{1}{c}{43.7} &
  \multicolumn{1}{c}{44.2} &
  \multicolumn{1}{c}{51.9} &
  \multicolumn{1}{c}{87.9} &
  \multicolumn{1}{c|}{\textbf{89.3}} &
  \multicolumn{1}{c}{44.1} &
  \multicolumn{1}{c}{46.3} &
  \multicolumn{1}{c}{51.4} &
  \multicolumn{1}{c}{\textbf{80.3}} &
  \multicolumn{1}{c|}{79.7} &
  \multicolumn{1}{c}{48.2} &
  \multicolumn{1}{c}{46.7} &
  \multicolumn{1}{c}{51.8} &
  \multicolumn{1}{c}{\textbf{81.8}} &
  81.1 \\ 
&
  4,000 &
  \multicolumn{1}{c}{43.9} &
  \multicolumn{1}{c}{45.7} &
  \multicolumn{1}{c}{56.5} &
  \multicolumn{1}{c}{91.4} &
  \multicolumn{1}{c|}{\textbf{91.6}} &
  \multicolumn{1}{c}{43.9} &
  \multicolumn{1}{c}{54.5} &
  \multicolumn{1}{c}{56.3} &
  \multicolumn{1}{c}{82.7} &
  \multicolumn{1}{c|}{\textbf{83.9}} &
  \multicolumn{1}{c}{46.3} &
  \multicolumn{1}{c}{49.2} &
  \multicolumn{1}{c}{56.3} &
  \multicolumn{1}{c}{\textbf{83.9}} &
  82.2 \\ 
&
  12,213 &
  \multicolumn{1}{c}{44.3} &
  \multicolumn{1}{c}{52.1} &
  \multicolumn{1}{c}{56.5} &
  \multicolumn{1}{c}{90.1} &
  \multicolumn{1}{c|}{\textbf{91.3}} &
  \multicolumn{1}{c}{44.0} &
  \multicolumn{1}{c}{50.7} &
  \multicolumn{1}{c}{56.4} &
  \multicolumn{1}{c}{82.1} &
  \multicolumn{1}{c|}{\textbf{83.7}} &
  \multicolumn{1}{c}{43.7} &
  \multicolumn{1}{c}{60.6} &
  \multicolumn{1}{c}{56.3} &
  \multicolumn{1}{c}{83.3} &
  \textbf{85.0} \\ 
  
\hline \hline 
\multicolumn{17}{c}{Source: \imi{} Sensor \textbf{1}} \\\hline
\hline

\multirow{4}{*}{2,000} & 50& \multicolumn{1}{c}{71.7} & \multicolumn{1}{c}{64.9}& \multicolumn{1}{c}{62.9} & \multicolumn{1}{c}{78} & \multicolumn{1}{c|}{\cellcolor[HTML]{FFFFFF}\textbf{79.4}}& \multicolumn{1}{c}{56.1}& \multicolumn{1}{c}{56.2} & \multicolumn{1}{c}{53.8}& \multicolumn{1}{c}{72.8} & \multicolumn{1}{c|}{\cellcolor[HTML]{FFFFFF}\textbf{75}} & \multicolumn{1}{c}{56.7}& \multicolumn{1}{c}{54.3} & \multicolumn{1}{c}{53.5} & \multicolumn{1}{c}{\cellcolor[HTML]{FFFFFF}\textbf{75.3}} & \multicolumn{1}{c|}{73.7}\\ 
                                                                                        & 500                                                                                      & \multicolumn{1}{c}{67}                                                                              & \multicolumn{1}{c}{69}                                                                             & \multicolumn{1}{c}{61.8}                                                                                   & \multicolumn{1}{c}{\cellcolor[HTML]{FFFFFF}\textbf{88.4}}                     & \multicolumn{1}{c|}{82.6}                                                                                                      & \multicolumn{1}{c}{56.1}                                                                            & \multicolumn{1}{c}{53.9}                                                                           & \multicolumn{1}{c}{56.5}                                                                                   & \multicolumn{1}{c}{\cellcolor[HTML]{FFFFFF}\textbf{78.7}}                     & \multicolumn{1}{c|}{\cellcolor[HTML]{FFFFFF}\textbf{78.7}}                                                                     & \multicolumn{1}{c}{56.6}                                                                            & \multicolumn{1}{c}{56.6}                                                                           & \multicolumn{1}{c}{56.7}                                                                                   & \multicolumn{1}{c}{79.4}                                                                                       & \cellcolor[HTML]{FFFFFF}\textbf{80.2}                                    \\ 
                                                                                      & 4,000                                                                                     & \multicolumn{1}{c}{71.8}                                                                            & \multicolumn{1}{c}{67.7}                                                                           & \multicolumn{1}{c}{56.2}                                                                                   & \multicolumn{1}{c}{\cellcolor[HTML]{FFFFFF}\textbf{90.6}}                     & \multicolumn{1}{c|}{88.5}                                                                                                      & \multicolumn{1}{c}{56.1}                                                                            & \multicolumn{1}{c}{56.3}                                                                           & \multicolumn{1}{c}{56.3}                                                                                   & \multicolumn{1}{c}{80.6}                                                      & \multicolumn{1}{c|}{\cellcolor[HTML]{FFFFFF}\textbf{81.6}}                                                                     & \multicolumn{1}{c}{56.7}                                                                            & \multicolumn{1}{c}{56.9}                                                                           & \multicolumn{1}{c}{57.3}                                                                                   & \multicolumn{1}{c}{81.2}                                                                                       & \cellcolor[HTML]{FFFFFF}\textbf{82.8}                                    \\ 
                                                                                        & 12,213                                                                                   & \multicolumn{1}{c}{73.6}                                                                            & \multicolumn{1}{c}{67.2}                                                                           & \multicolumn{1}{c}{56.3}                                                                                   & \multicolumn{1}{c}{90.5}                                                      & \multicolumn{1}{c|}{\cellcolor[HTML]{FFFFFF}\textbf{91.5}}                                                                     & \multicolumn{1}{c}{56}                                                                              & \multicolumn{1}{c}{56.3}                                                                           & \multicolumn{1}{c}{56.2}                                                                                   & \multicolumn{1}{c}{78.9}                                                      & \multicolumn{1}{c|}{\cellcolor[HTML]{FFFFFF}\textbf{82.7}}                                                                     & \multicolumn{1}{c}{56.6}                                                                            & \multicolumn{1}{c}{56.4}                                                                           & \multicolumn{1}{c}{56.3}                                                                                   & \multicolumn{1}{c}{80.5}                                                                                       & \cellcolor[HTML]{FFFFFF}\textbf{84.2}                                    \\ 
\hline
\multirow{4}{*}{8,000}                                                                                        & 50                                                                                       & \multicolumn{1}{c}{71.8}                                                                            & \multicolumn{1}{c}{70.9}                                                                           & \multicolumn{1}{c}{50.5}                                                                                   & \multicolumn{1}{c}{77.8}                                                      & \multicolumn{1}{c|}{\cellcolor[HTML]{FFFFFF}\textbf{79}}                                                                       & \multicolumn{1}{c}{56.4}                                                                            & \multicolumn{1}{c}{56.3}                                                                           & \multicolumn{1}{c}{56.2}                                                                                   & \multicolumn{1}{c}{70.6}                                                      & \multicolumn{1}{c|}{\cellcolor[HTML]{FFFFFF}\textbf{72.2}}                                                                     & \multicolumn{1}{c}{56.6}                                                                            & \multicolumn{1}{c}{55.8}                                                                           & \multicolumn{1}{c}{54.1}                                                                                   & \multicolumn{1}{c}{71.5}                                                                                       & \cellcolor[HTML]{FFFFFF}\textbf{73.7}                                    \\ 
                                                                                        & 500                                                                                      & \multicolumn{1}{c}{60.8}                                                                            & \multicolumn{1}{c}{64.4}                                                                           & \multicolumn{1}{c}{56.2}                                                                                   & \multicolumn{1}{c}{\cellcolor[HTML]{FFFFFF}\textbf{88.9}}                     & \multicolumn{1}{c|}{88.8}                                                                                                      & \multicolumn{1}{c}{56.2}                                                                            & \multicolumn{1}{c}{57}                                                                             & \multicolumn{1}{c}{56.3}                                                                                   & \multicolumn{1}{c}{\cellcolor[HTML]{FFFFFF}\textbf{80.4}}                     & \multicolumn{1}{c|}{80.3}                                                                                                      & \multicolumn{1}{c}{56.6}                                                                            & \multicolumn{1}{c}{56.1}                                                                           & \multicolumn{1}{c}{53.9}                                                                                   & \multicolumn{1}{c}{81.1}                                                                                       & \cellcolor[HTML]{FFFFFF}\textbf{81.4}                                    \\ 
                                                                                        & 4,000                                                                                     & \multicolumn{1}{c}{61}                                                                              & \multicolumn{1}{c}{80.6}                                                                           & \multicolumn{1}{c}{56.3}                                                                                   & \multicolumn{1}{c}{89.5}                                                      & \multicolumn{1}{c|}{\cellcolor[HTML]{FFFFFF}\textbf{91.5}}                                                                     & \multicolumn{1}{c}{56.1}                                                                            & \multicolumn{1}{c}{56.3}                                                                           & \multicolumn{1}{c}{56.3}                                                                                   & \multicolumn{1}{c}{76.6}                                                      & \multicolumn{1}{c|}{\cellcolor[HTML]{FFFFFF}\textbf{82.7}}                                                                     & \multicolumn{1}{c}{56.5}                                                                            & \multicolumn{1}{c}{56.5}                                                                           & \multicolumn{1}{c}{56.3}                                                                                   & \multicolumn{1}{c}{83.1}                                                                                       & \cellcolor[HTML]{FFFFFF}\textbf{84.5}                                    \\ 
                                                                                       & 12,213                                                                                    & \multicolumn{1}{c}{67}                                                                              & \multicolumn{1}{c}{67.1}                                                                           & \multicolumn{1}{c}{56.3}                                                                                   & \multicolumn{1}{c}{90.7}                                                      & \multicolumn{1}{c|}{\cellcolor[HTML]{FFFFFF}\textbf{91.3}}                                                                     & \multicolumn{1}{c}{56.2}                                                                            & \multicolumn{1}{c}{56.4}                                                                           & \multicolumn{1}{c}{56.2}                                                                                   & \multicolumn{1}{c}{80.1}                                                      & \multicolumn{1}{c|}{\cellcolor[HTML]{FFFFFF}\textbf{83.8}}                                                                     & \multicolumn{1}{c}{56.6}                                                                            & \multicolumn{1}{c}{56.7}                                                                           & \multicolumn{1}{c}{56.3}                                                                                   & \multicolumn{1}{c}{81.6}                                                                                       & \cellcolor[HTML]{FFFFFF}\textbf{84.2}                                    \\ 
\hline
\multirow{4}{*}{16,000}                                                                                      & 50                                                                                       & \multicolumn{1}{c}{62.7}                                                                            & \multicolumn{1}{c}{68.5}                                                                           & \multicolumn{1}{c}{56.2}                                                                                   & \multicolumn{1}{c}{\cellcolor[HTML]{FFFFFF}\textbf{80.8}}                     & \multicolumn{1}{c|}{78.7}                                                                                                      & \multicolumn{1}{c}{56.1}                                                                            & \multicolumn{1}{c}{56.1}                                                                           & \multicolumn{1}{c}{54.2}                                                                                   & \multicolumn{1}{c}{\cellcolor[HTML]{FFFFFF}\textbf{72.6}}                     & \multicolumn{1}{c|}{72.5}                                                                                                      & \multicolumn{1}{c}{56.1}                                                                            & \multicolumn{1}{c}{56.6}                                                                           & \multicolumn{1}{c}{53.9}                                                                                   & \multicolumn{1}{c}{\cellcolor[HTML]{FFFFFF}\textbf{73.8}}                                                      & 73.4                                                                     \\ 
                                                                                      & 500                                                                                      & \multicolumn{1}{c}{67.2}                                                                            & \multicolumn{1}{c}{62.8}                                                                           & \multicolumn{1}{c}{56.2}                                                                                   & \multicolumn{1}{c}{87.5}                                                      & \multicolumn{1}{c|}{\cellcolor[HTML]{FFFFFF}\textbf{88.7}}                                                                     & \multicolumn{1}{c}{56.2}                                                                            & \multicolumn{1}{c}{55.9}                                                                           & \multicolumn{1}{c}{56.3}                                                                                   & \multicolumn{1}{c}{\cellcolor[HTML]{FFFFFF}\textbf{80.5}}                     & \multicolumn{1}{c|}{77.4}                                                                                                      & \multicolumn{1}{c}{56.6}                                                                            & \multicolumn{1}{c}{54}                                                                             & \multicolumn{1}{c}{53.8}                                                                                   & \multicolumn{1}{c}{81.8}                                                                                       & \cellcolor[HTML]{FFFFFF}\textbf{82.3}                                    \\ 
& 4,000                                                                                     & \multicolumn{1}{c}{67.6}                                                                            & \multicolumn{1}{c}{79.6}                                                                           & \multicolumn{1}{c}{56.3}                                                                                   & \multicolumn{1}{c}{90.9}                                                      & \multicolumn{1}{c|}{\cellcolor[HTML]{FFFFFF}\textbf{92.2}}                                                                     & \multicolumn{1}{c}{56.2}                                                                            & \multicolumn{1}{c}{56.5}                                                                           & \multicolumn{1}{c}{56.3}                                                                                   & \multicolumn{1}{c}{82.2}                                                      & \multicolumn{1}{c|}{\cellcolor[HTML]{FFFFFF}\textbf{84.2}}                                                                     & \multicolumn{1}{c}{55.6}                                                                            & \multicolumn{1}{c}{56.3}                                                                           & \multicolumn{1}{c}{56.3}                                                                                   & \multicolumn{1}{c}{84}                                                                                         & \cellcolor[HTML]{FFFFFF}\textbf{85.1}                                    \\ 
& 12,213                                                                                    & \multicolumn{1}{c}{59.7}                                                                            & \multicolumn{1}{c}{79.1}                                                                           & \multicolumn{1}{c}{56.3}                                                                                   & \multicolumn{1}{c}{90.5}                                                      & \multicolumn{1}{c|}{\cellcolor[HTML]{FFFFFF}\textbf{91.4}}                                                                     & \multicolumn{1}{c}{56.1}                                                                            & \multicolumn{1}{c}{56.2}                                                                           & \multicolumn{1}{c}{56.3}                                                                                   & \multicolumn{1}{c}{81.8}                                                      & \multicolumn{1}{c|}{\cellcolor[HTML]{FFFFFF}\textbf{83.6}}                                                                     & \multicolumn{1}{c}{56.6}                                                                            & \multicolumn{1}{c}{56.6}                                                                           & \multicolumn{1}{c}{56.3}                                                                                   & \multicolumn{1}{c}{83.2}                                                                                       & \cellcolor[HTML]{FFFFFF}\textbf{84.9}                                    \\

\hline \hline 
\multicolumn{17}{c}{Source: \imi{} Sensor \textbf{2}} \\\hline
\hline

\multirow{4}{*}{2,000} & 50& \multicolumn{1}{c}{49.5} & \multicolumn{1}{c}{53.7} & \multicolumn{1}{c}{50.3} & \multicolumn{1}{c}{\cellcolor[HTML]{FFFFFF}\textbf{79.8}} & \multicolumn{1}{c|}{79.7} & \multicolumn{1}{c}{58.4} & \multicolumn{1}{c}{57.6} & \multicolumn{1}{c}{52.5} & \multicolumn{1}{c}{71.6} & \multicolumn{1}{c|}{\cellcolor[HTML]{FFFFFF}\textbf{72.9}} & \multicolumn{1}{c}{52.7} & \multicolumn{1}{c}{52.3} & \multicolumn{1}{c}{53.1} & \multicolumn{1}{c}{74.0} & \multicolumn{1}{c|}{\cellcolor[HTML]{FFFFFF}\textbf{74.1}}\\ 

& 500                                                                                      & \multicolumn{1}{c}{45.4}                                                                            & \multicolumn{1}{c}{57.3}                                                                           & \multicolumn{1}{c}{55.7}                                                                                   & \multicolumn{1}{c}{\cellcolor[HTML]{FFFFFF}\textbf{88.6}}                     & \multicolumn{1}{c|}{88.5}                                                                     & \multicolumn{1}{c}{58.1}                                                                            & \multicolumn{1}{c}{56.4}                                                                           & \multicolumn{1}{c}{55.2}                                                                                   & \multicolumn{1}{c}{78.6}                                                      & \multicolumn{1}{c|}{\cellcolor[HTML]{FFFFFF}\textbf{79.3}}                                    & \multicolumn{1}{c}{51.6}                                                                            & \multicolumn{1}{c}{58.6}                                                                           & \multicolumn{1}{c}{58.7}                                                                                   & \multicolumn{1}{c}{78.9}                                                      & \cellcolor[HTML]{FFFFFF}\textbf{79.3}                                    \\ 
& 4,000                                                                                     & \multicolumn{1}{c}{47}                                                                              & \multicolumn{1}{c}{56.3}                                                                           & \multicolumn{1}{c}{56.3}                                                                                   & \multicolumn{1}{c}{\cellcolor[HTML]{FFFFFF}\textbf{89.2}}                     & \multicolumn{1}{c|}{89.1}                                                                     & \multicolumn{1}{c}{56.7}                                                                            & \multicolumn{1}{c}{56.3}                                                                           & \multicolumn{1}{c}{56.3}                                                                                   & \multicolumn{1}{c}{80.6}                                                      & \multicolumn{1}{c|}{\cellcolor[HTML]{FFFFFF}\textbf{81.6}}                                    & \multicolumn{1}{c}{52.6}                                                                            & \multicolumn{1}{c}{56.3}                                                                           & \multicolumn{1}{c}{56.3}                                                                                   & \multicolumn{1}{c}{80.9}                                                      & \cellcolor[HTML]{FFFFFF}\textbf{82.7}                                    \\ 
& 12,213                                                                                    & \multicolumn{1}{c}{45.8}                                                                            & \multicolumn{1}{c}{56.3}                                                                           & \multicolumn{1}{c}{56.3}                                                                                   & \multicolumn{1}{c}{88.5}                                                      & \multicolumn{1}{c|}{\cellcolor[HTML]{FFFFFF}\textbf{91.7}}                                    & \multicolumn{1}{c}{58.5}                                                                            & \multicolumn{1}{c}{56.3}                                                                           & \multicolumn{1}{c}{56.3}                                                                                   & \multicolumn{1}{c}{81.4}                                                      & \multicolumn{1}{c|}{\cellcolor[HTML]{FFFFFF}\textbf{82.9}}                                    & \multicolumn{1}{c}{53.2}                                                                            & \multicolumn{1}{c}{56.3}                                                                           & \multicolumn{1}{c}{56.3}                                                                                   & \multicolumn{1}{c}{81.3}                                                      & \cellcolor[HTML]{FFFFFF}\textbf{84.3}                                    \\ 
\hline \multirow{4}{*}{8,000}                                                                                        & 50                                                                                       & \multicolumn{1}{c}{48.7}                                                                            & \multicolumn{1}{c}{52.7}                                                                           & \multicolumn{1}{c}{51.5}                                                                                   & \multicolumn{1}{c}{\cellcolor[HTML]{FFFFFF}\textbf{80}}                       & \multicolumn{1}{c|}{77.3}                                                                     & \multicolumn{1}{c}{56.2}                                                                            & \multicolumn{1}{c}{56.6}                                                                           & \multicolumn{1}{c}{53.8}                                                                                   & \multicolumn{1}{c}{72.8}                                                      & \multicolumn{1}{c|}{\cellcolor[HTML]{FFFFFF}\textbf{74.4}}                                    & \multicolumn{1}{c}{51.2}                                                                            & \multicolumn{1}{c}{54.2}                                                                           & \multicolumn{1}{c}{51.5}                                                                                   & \multicolumn{1}{c}{72.9}                                                      & \cellcolor[HTML]{FFFFFF}\textbf{73.6}                                    \\ 
& 500                                                                                      & \multicolumn{1}{c}{56.9}                                                                            & \multicolumn{1}{c}{52.6}                                                                           & \multicolumn{1}{c}{55.8}                                                                                   & \multicolumn{1}{c}{88.5}                                                      & \multicolumn{1}{c|}{\cellcolor[HTML]{FFFFFF}\textbf{88.8}}                                    & \multicolumn{1}{c}{57.8}                                                                            & \multicolumn{1}{c}{57.9}                                                                           & \multicolumn{1}{c}{53.9}                                                                                   & \multicolumn{1}{c}{\cellcolor[HTML]{FFFFFF}\textbf{79.9}}                     & \multicolumn{1}{c|}{79.8}                                                                     & \multicolumn{1}{c}{52.1}                                                                            & \multicolumn{1}{c}{57.6}                                                                           & \multicolumn{1}{c}{56.7}                                                                                   & \multicolumn{1}{c}{81.4}                                                      & \cellcolor[HTML]{FFFFFF}\textbf{82.3}                                    \\ 
& 4,000                                                                                     & \multicolumn{1}{c}{50.6}                                                                            & \multicolumn{1}{c}{59.8}                                                                           & \multicolumn{1}{c}{56.3}                                                                                   & \multicolumn{1}{c}{90.6}                                                      & \multicolumn{1}{c|}{\cellcolor[HTML]{FFFFFF}\textbf{91}}                                      & \multicolumn{1}{c}{57.7}                                                                            & \multicolumn{1}{c}{56.3}                                                                           & \multicolumn{1}{c}{56.3}                                                                                   & \multicolumn{1}{c}{81.3}                                                      & \multicolumn{1}{c|}{\cellcolor[HTML]{FFFFFF}\textbf{83.3}}                                    & \multicolumn{1}{c}{54.3}                                                                            & \multicolumn{1}{c}{56.3}                                                                           & \multicolumn{1}{c}{56.3}                                                                                   & \multicolumn{1}{c}{83.1}                                                      & \cellcolor[HTML]{FFFFFF}\textbf{84.2}                                    \\ 
                                                                                        & 12,213                                                                                    & \multicolumn{1}{c}{49.9}                                                                            & \multicolumn{1}{c}{63.3}                                                                           & \multicolumn{1}{c}{56.5}                                                                                   & \multicolumn{1}{c}{88}                                                        & \multicolumn{1}{c|}{\cellcolor[HTML]{FFFFFF}\textbf{91.3}}                                    & \multicolumn{1}{c}{56.9}                                                                            & \multicolumn{1}{c}{56}                                                                             & \multicolumn{1}{c}{56.3}                                                                                   & \multicolumn{1}{c}{80.6}                                                      & \multicolumn{1}{c|}{\cellcolor[HTML]{FFFFFF}\textbf{82.5}}                                    & \multicolumn{1}{c}{53.6}                                                                            & \multicolumn{1}{c}{56.3}                                                                           & \multicolumn{1}{c}{56.5}                                                                                   & \multicolumn{1}{c}{81.9}                                                      & \cellcolor[HTML]{FFFFFF}\textbf{84.3}                                    \\ \hline
\multirow{4}{*}{16,000}    & 50                                                                                       & \multicolumn{1}{c}{51.2}                                                                            & \multicolumn{1}{c}{55.1}                                                                           & \multicolumn{1}{c}{51.4}                                                                                   & \multicolumn{1}{c}{75.1}                                                      & \multicolumn{1}{c|}{\cellcolor[HTML]{FFFFFF}\textbf{78.9}}                                    & \multicolumn{1}{c}{54.7}                                                                            & \multicolumn{1}{c}{57.2}                                                                           & \multicolumn{1}{c}{51.7}                                                                                   & \multicolumn{1}{c}{\cellcolor[HTML]{FFFFFF}\textbf{72.8}}                     & \multicolumn{1}{c|}{71.6}                                                                     & \multicolumn{1}{c}{52.1}                                                                            & \multicolumn{1}{c}{54.3}                                                                           & \multicolumn{1}{c}{46.4}                                                                                   & \multicolumn{1}{c}{72.3}                                                      & \cellcolor[HTML]{FFFFFF}\textbf{73.9}                                    \\ 
& 500                                                                                      & \multicolumn{1}{c}{47.9}                                                                            & \multicolumn{1}{c}{56.3}                                                                           & \multicolumn{1}{c}{56.4}                                                                                   & \multicolumn{1}{c}{88.1}                                                      & \multicolumn{1}{c|}{\cellcolor[HTML]{FFFFFF}\textbf{88.7}}                                    & \multicolumn{1}{c}{57.6}                                                                            & \multicolumn{1}{c}{60.1}                                                                           & \multicolumn{1}{c}{56.3}                                                                                   & \multicolumn{1}{c}{\cellcolor[HTML]{FFFFFF}\textbf{81}}                       & \multicolumn{1}{c|}{79.5}                                                                     & \multicolumn{1}{c}{49.4}                                                                            & \multicolumn{1}{c}{52.5}                                                                           & \multicolumn{1}{c}{56.5}                                                                                   & \multicolumn{1}{c}{81.7}                                                      & \cellcolor[HTML]{FFFFFF}\textbf{81.8}                                    \\ 
& 4,000                                                                                     & \multicolumn{1}{c}{51.6}                                                                            & \multicolumn{1}{c}{65.1}                                                                           & \multicolumn{1}{c}{56.3}                                                                                   & \multicolumn{1}{c}{90.5}                                                      & \multicolumn{1}{c|}{\cellcolor[HTML]{FFFFFF}\textbf{91}}                                      & \multicolumn{1}{c}{58.3}                                                                            & \multicolumn{1}{c}{59.1}                                                                           & \multicolumn{1}{c}{56.1}                                                                                   & \multicolumn{1}{c}{83}                                                        & \multicolumn{1}{c|}{\cellcolor[HTML]{FFFFFF}\textbf{84}}                                      & \multicolumn{1}{c}{52.4}                                                                            & \multicolumn{1}{c}{60.3}                                                                           & \multicolumn{1}{c}{56.3}                                                                                   & \multicolumn{1}{c}{82.9}                                                      & \cellcolor[HTML]{FFFFFF}\textbf{84.7}                                    \\ 
& 12,213                                                                                    & \multicolumn{1}{c}{47.3}                                                                            & \multicolumn{1}{c}{58.4}                                                                           & \multicolumn{1}{c}{56.3}                                                                                   & \multicolumn{1}{c}{90.4}                                                      & \multicolumn{1}{c|}{\cellcolor[HTML]{FFFFFF}\textbf{91}}                                      & \multicolumn{1}{c}{56.8}                                                                            & \multicolumn{1}{c}{56.3}                                                                           & \multicolumn{1}{c}{56.3}                                                                                   & \multicolumn{1}{c}{82.1}                                                      & \multicolumn{1}{c|}{\cellcolor[HTML]{FFFFFF}\textbf{83.5}}                                    & \multicolumn{1}{c}{53.4}                                                                            & \multicolumn{1}{c}{59.8}                                                                           & \multicolumn{1}{c}{56.3}                                                                                   & \multicolumn{1}{c}{82.8}                                                      & \cellcolor[HTML]{FFFFFF}\textbf{85}                                      \\ \hline
\hline
\multicolumn{17}{c}{Statistics on Target Test Accuracy of models on different target sensors } \\\hline
\hline 
\multicolumn{2}{|c|}{Min}    & 43.4 & 40.7 & 46.1 & 75.1          &    \multicolumn{1}{c|}{\textbf{77.3}} & 43.3 & 43.9 & 44.4 & 70.6 & \multicolumn{1}{c|}{\textbf{71.6}} & 43.3 & 43.8 & 46.4 & 71.5 & \textbf{72}   \\ 
\multicolumn{2}{|c|}{Median} & 49.1 & 56.3 & 56.3 & 88.6          & \multicolumn{1}{c|}{\textbf{89.1}} & 56.1 & 56.3 & 56.3 & 80.3 & \multicolumn{1}{c|}{\textbf{81.1}} & 52.5 & 56.3 & 56.3 & 81.2 & \textbf{82.3} \\ 
\multicolumn{2}{|c|}{Max}    & 73.6 & 80.6 & 62.9 & \textbf{92.5} & \multicolumn{1}{c|}{92.2}          & 58.5 & 60.1 & 56.5 & 83   & \multicolumn{1}{c|}{\textbf{84.2}} & 56.7 & 60.6 & 58.7 & 84   & \textbf{85.1} \\ 
\multicolumn{2}{|c|}{Mean}   & 53.6 & 58.3 & 55   & 87            & \multicolumn{1}{c|}{\textbf{87.3}} & 52.5 & 54.1 & 54.7 & 78.5 & \multicolumn{1}{c|}{\textbf{79.7}} & 51.1 & 54.1 & 54.5 & 79.6 & \textbf{80.8} \\ 
\hline
\end{tabular}

\end{table*}

\subsection{RQ2 Experiment}
In this experiment, we compared our adversarial DA models with baseline models, for classifying unlabeled target data by learning on large labeled source domain data and small labeled target domain data. We ran experiments on nine combinations: from all three sensors of the source domain (\imi{}) to all three sensors of domain (\kirby{}). We compared five models: three vanilla baselines (single, multi and fine-tuned), and two adversarial DA models (\icit{}, \icitmod{}). For brevity, we refer to the \textbf{Vanilla} \textbf{B}aseline \textbf{S}ingle, \textbf{M}ulti and \textbf{F}ine Tuned \textbf{M}odels respectively as \vbs{}, \vbm{} and \vbf{}. 

All the vanilla baselines are \textbf{MLP}s with two hidden layers of 512 and 200 nodes. Both of these have ReLU as activation, and dropout layers with $p=0.2$. In final layer, Softmax is applied. The difference among them is in how they are trained: \vbs{} is trained and validated with source domain data; \vbm{} is trained and validated with both source and target domain data; and finally, \vbf{} is at first trained and validated on source domain data, and later, the last dense and output layers are fine-tuned with target domain data, while the first dense layer weights remain frozen. For the vanilla baselines, we have used cross-entropy loss with the Adam optimizer \cite{kingma2014adam}, with learning rate 0.001 and $\beta_1 =$ 0.9.

For the \icit{}  and \icitmod{} models, we train and validate on both source and target domain (train and validation split respectively) data, using Adam optimizer~\cite{kingma2014adam} for all components with the same learning rate (=0.0002) and $\beta_1$(=0.5). The values of loss weights $\beta, \gamma$, and $\lambda$ are all set to 1 after trial and error over different ranges.   
Batch size and learning epoch were set to 128 and 20, respectively. Moreover, for \icit{} \& \icitmod{}, in each epoch, we iterated over the entire training data for both domains at least once. 
In each epoch, after a fixed number of batch steps (=40) iteratively, we performed validation and saved the model with best target validation accuracy. We iterated over the pairing of source training data size = 
\{
2000, 
8,000, 
16,000\} 
and target training data size = 
\{50, 
500, 
4,000, 
12,213\} 
and ran each combination with all models, for three times with fixed seeds, and reported the average accuracy. For the model architectures, refer to the \Cref{sec:mod:arch}. 

\begin{figure}
    \centering
    \includegraphics[width=0.85\linewidth]{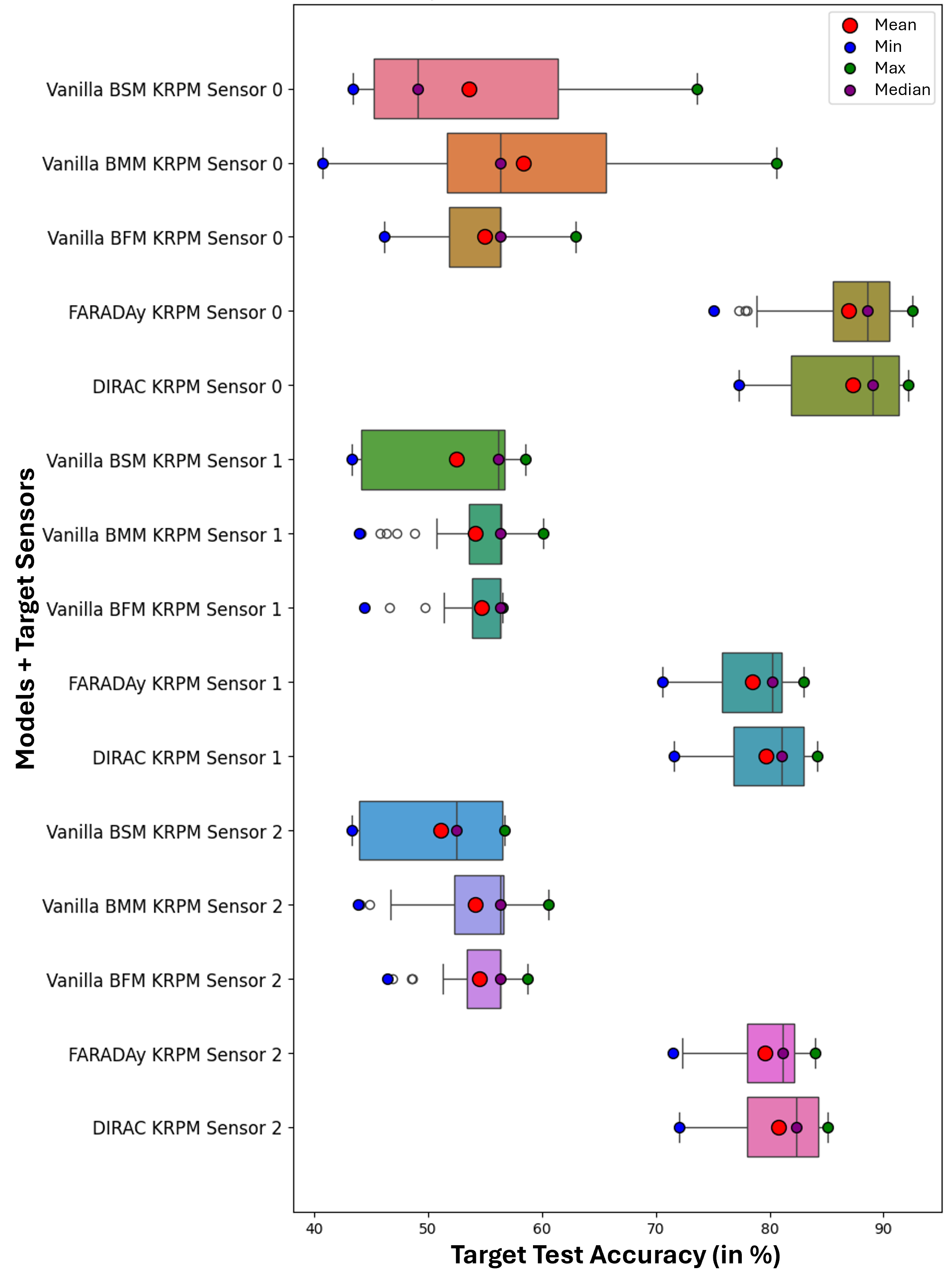}
    \caption{Boxplot for Target Domain Test accuracy for Transfer Learning Models with \kirby{} Sensors as Target }
    \label{fig:boxplot}
\end{figure}

\Cref{tab:rq2sensall_reduced_new} shows the performances of the different models, 
from \imi{} Sensor 0, 1, and 2 respectively, to all \kirby{} sensors. 
From the results, we can see that: \textbf{(1)} \icit{} and \icitmod{} outperform all the vanilla baseline models by a considerable margin. From the statistics shown in \Cref{tab:rq2sensall_reduced_new}, we can see that for \kirby{} Sensor 0, the mean value of the target test accuracies of our models (87\%) are higher than even the maximum value attained by the baseline models (80.6\%). For Sensor 1 and 2, the minimum values of our models (70.6 \%, 71.5\%) are higher than the maximum values attained by the baseline models (60.1\%, 60.6\%). These distributions of the accuracies are shown in the boxplot in \Cref{fig:boxplot}. \textbf{(2)} For each data size combination- among all sensors, our models showed their best performances for \kirby{} Sensor 0 data. For example, for \imi{} Sensor 1 as source with source size 8000, and target size 4000- the best performance among all models and target sensor combinations is achieved by \icitmod{}- 91.5\% for \kirby{} Sensor 0, whereas for target size 500 it is achieved by \icit{} (88.9\%), also in Sensor 0.  This indicates that \kirby{} Sensor 0, the very sensor in the closest proximity to the cutting machine, captures the sound information best. The statistics shown in the \Cref{tab:rq2sensall_reduced_new} also prove the superiority of the Sensor 0 data capturing ability, as in each metric Sensor 0 is ahead of Sensor 1 and 2. 
. These findings provide valuable guidelines for sensor selection in industrial monitoring implementation, suggesting that placing sensors in close proximity to the cutting machine, can significantly enhance the accuracy of sound-based inferences.

\subsection{The qualitative evaluation and reason of \icit{} \& \icitmod{}'s effectiveness}
\begin{figure}
    \centering
   \begin{subfigure}{\linewidth}
    \centering
    \includegraphics[width=0.82\linewidth]{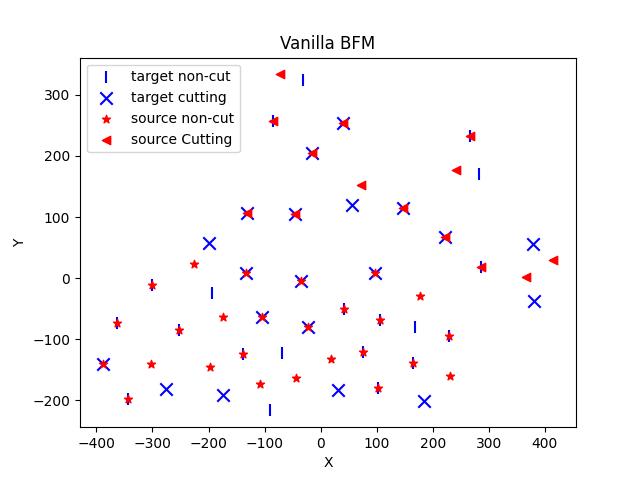}
    \caption{t-SNE Projection of \textbf{Vanilla BFM}}
    \label{fig:tsnevanbfm}
    \end{subfigure}
\begin{subfigure}{\linewidth}
    \centering
    \includegraphics[width=0.82\linewidth]{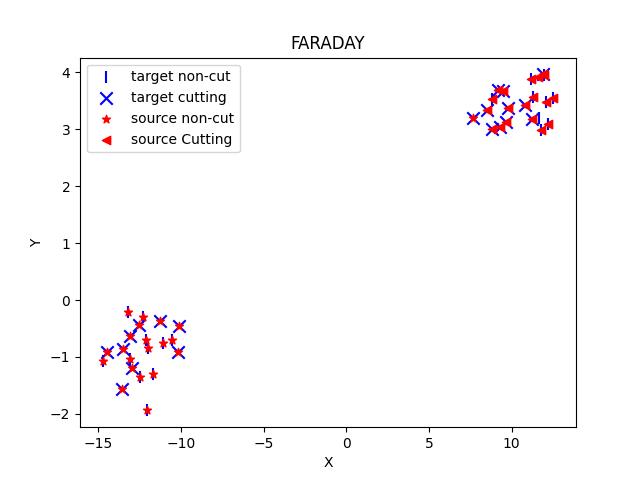}
    \caption{t-SNE Projection of \icit{}}
    \label{fig:tsnedirac}

\end{subfigure}
\begin{subfigure}{\linewidth}
    \centering
    \includegraphics[width=0.82\linewidth]{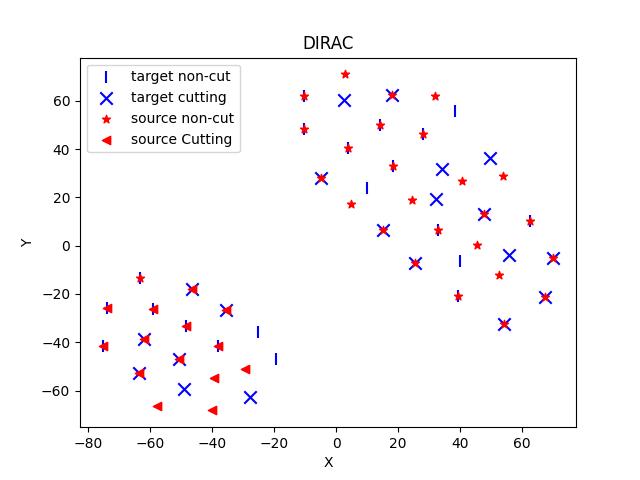}
    \caption{t-SNE Projection of \icitmod{}}
    \label{fig:tsnefaraday}
\end{subfigure}
\end{figure}

As a demonstration of why our methods are more effective than Vanilla methods, we conduct a small experiment. We choose 2000 \imi{} source samples and 1000 \kirby{} target samples, both from sensor 0, and ran \textbf{Vanilla BFM}, \icit{} \& \icitmod{} on them. Later, we projected the latent representation of the samples (shared generator/ dense layer outputs) on 2- dimensional space using t-SNE (\cite{hinton2002stochastic}) algorithm in \Cref{fig:tsnevanbfm,fig:tsnefaraday,fig:tsnedirac} (only 40 source and 40 target samples shown in figures). We see that the decision boundaries for both domains and cuttings are inseparable for \textbf{Vanilla BFM}, whereas the DA models have clustered them efficiently. \icit{} and \icitmod{} both have successfully separated the cutting and non-cut source data, whereas for the target there are some outliers. Nevertheless, most of the target non-cut and cutting data are aligned respectively with source non-cut and cutting data, proving domain-invariance learning of our methods. Also, \icitmod{} clusters are more spread than \icit{} ones, as domain information is ingrained there.

\section{Discussion}
From the results in \Cref{sec:exp}, we have seen that \icit{} and \icitmod{} achieve near-optimal results while learning from both lab data and a few industry data, and inferring on industry data. This matches with our hypothesis that if we separate domain-dependent and domain-independent features leveraging adversarial learning, we can learn a shared feature space which is used to classify data from both source and target domain without domain labels. These results also demonstrated effective domain adaptation across different types and placements of sound sensors. 
Furthermore, transforming the features before running the models helps reduce the training load as no additional layers were needed to transform the feature spaces into a common range. 
Nevertheless, the model still has some limitations. First, our GPU limitations let us only handle 16,000 and 12,213 training samples from both domains. To learn a better representation, we need to find out a way to load in memory all 261,560 samples from the dataset. Second, our current model can only be applied to data of the same type, that is, sound, image or text.
We have not yet tested our model on multi-modal data, which involves training and testing samples that contain different types of data. Further testing is needed to determine if our methodology requires extensions.


\section{Conclusion}
In this paper, we proposed an adversarial domain adaptation approach that utilizes abundant lab data along with scarce industry data to train a model capable of inferring the cutting state of milling processes from audible sound. We collected and published cutting sound datasets from both lab and industry settings, using multiple sensors in different locations. These datasets were then used to evaluate the proposed approach. We analyzed the impact of different sound feature transformations based on FFT on the model's accuracy. Overall, the Gamma transformation yielded the best training accuracy when applied to a single target domain. Using the same feature transformation, we evaluated the transfer learning performance of the proposed adversarial domain adaptation approach with the augmented features compared to the vanilla baseline models across various source and target data sizes. Experimental results showed that our approach outperformed other baseline methods across different types and locations of the sensors. Specifically, the prediction accuracies were approximately 92\% for Sensor 1 (microphone near cutting area), 82\% for Sensor 2 (ISS at spindle), and 85\% for Sensor 3 (ISS at base) in the various combinations of the target and source sensors. This indicates that our approach can be effectively applied even in scenarios where sensor selection and installation locations are limited in industrial settings.

Several extensions and improvements are possible. 
Our model could be extended for multi-modal learning on various types of data beyond sound. Besides, cross-sensor learning from sound data could enhance the model's robustness.
Another interesting direction is to explore the model's performance with few or no labeled data from the target domain during training.
\section*{Acknowledgement}
This research was supported by NSF Grant No. 2134667 "FMRG: Manufacturing USA: Cyber: Privacy-Preserving Tiny Machine Learning Edge Analytics to Enable AI-Commons for Secure Manufacturing".

\bibliographystyle{IEEEtran}
\bibliography{main}

\end{document}